# Feature selection using nearest attributes

Alex Pappachen James, *Member, IEEE,* and Sima Dimitrijev, *Senior Member, IEEE*

**Abstract**—Feature selection is an important problem in high-dimensional data analysis and classification. Conventional feature selection approaches focus on detecting the features based on a redundancy criterion using learning and feature searching schemes. In contrast, we present an approach that identifies the need to select features based on their discriminatory ability among classes. Area of overlap between inter-class and intra-class distances resulting from feature to feature comparison of an attribute is used as a measure of discriminatory ability of the feature. A set of nearest attributes in a pattern having the lowest area of overlap within a degree of tolerance defined by a selection threshold is selected to represent the best available discriminable features. State of the art recognition results are reported for pattern classification problems by using the proposed feature selection scheme with the nearest neighbour classifier. These results are reported with benchmark databases having high dimensional feature vectors in the problems involving images and micro array data.

✦

## 1 INTRODUCTION

HIGH resolution images, Internet web, financial monitoring, and DNA micro-arrays are some of the sources for databases used in pattern recognition research [1]–[15]. Many of these contemporary databases uses considerably large number of data points to represent an object sample. High dimensional feature vectors that result from these samples often contain intra-class natural variability reflected as noise and irrelevant information [1]–[10]. The noise in feature vectors occurs due to inaccurate feature measurements, whereas irrelevancy of a feature depends on the natural variability and the redundancy within the feature vector. Further, relevance of a feature is application depended; for example, consider a hypothetical image which has images of faces and objects. When using this image in a face recognition application, the relevant pixels in the image are in face regions while rest of the regions are considered as irrelevant. In addition, face region itself can have irrelevant information due to presence of intra-class variability such as occlusions, facial expressions, illumination changes and pose changes. Natural variability that occurs in high dimensional data has significant impact on lowering the performance of all pattern recognition methods. To improve the performance of pattern recognition methods, in the past, most of the effort has been to compensate or remove intra-class natural variability from the data samples through various feature processing methods.

- *A. P. James is a A/ Professor with Intelligent Machines and Neuromorphic Engineering Laboratory, IIITM-Kerala, apj@ieee.org.*
- *S. Dimitrijev is a Professor at Griffith School of Engineering, Griffith University and is the Deputy Director of Queensland Micro- and Nanotechnology centre.*

Dimensionality reduction [16]–[19] and feature selection [11]–[15], [20] are two types of feature processing techniques that are used to automatically improve the quality of data by removing irrelevant information. Dimensionality reduction methods are popular as it achieves the purpose of reducing the number of features and noise in a feature vector at the mathematical convenience of feature transformations and projections. However, the assumption of correlations between the features in the data is a core aspect of dimensionality reduction methods that can result in inaccurate feature descriptions. Further, irrelevant information from the original data is not always possible to remove in a dimensionality reduction approach. The problem to improve the quality of resulting features using linear and more recently non-linear dimensionality reduction methods has consistently been a field of intense research and debate in the recent past [16]. An alternative to dimensionality reduction approach to improve feature quality is by removing the irrelevant features from the high dimensional feature vector using feature selection methods. Feature selection methods have been an intense field of study in the recent few years and has gained its importance in parallel to most dimensionality reduction methods. Feature selection provides an undue advantage over dimensionality reduction methods for its ability to distinguish and select the best available features in a data [11]–[15], [20]. This would mean that feature selection methods can be applied without losing its generality to original feature vectors and also to those feature vectors that result from the application of dimensionality reduction methods. With this point of view, feature selection can be considered as an essential component required for developing high performance pattern classification systems that uses high dimensional data [1]–[10]. Since higher dimensional feature vectors contain several irrelevant features that reduce the performance of pattern recognition methods, feature selection by itself can be used in most of the modern pattern recognition methods to combat the issues resulting from the curse of high dimensionality [1]–[10]. There are two main types of feature selection methods: (1) that uses learning techniques [21]–[28], and (2) that uses criteria/filter based feature searching [29]–[31]. Features are selected based on the rank as obtained by evaluating individual features against a selection criterion such that redundancy of features in the dataset is minimised. Most of the methods rely on learning strategies including feature relevance calculations to select features holistically [21]–[28]. Learning based solutions results in features that tend to be more responsive to the changes in training data, and the feature selection largely depends on the feature relevance within a class [21]–[28]. However, an increase in dimensionality makes the learning algorithms computationally intensive and also results in the need for the development of various optimization techniques. Criteria driven methods are computationally less complex, and robust to changes in training data [29]–[31]. In such methods, the main idea is to optimize an objective function using a common approach of forward or backward search of the features. In such approaches the ability to accurately determine the most important features depend heavily on the objective function. Variations in the nature of data from a database to another makes the optimal selection of objective function difficult and a high classification accuracy using selected features from such methods are not always guaranteed.

Feature selection by itself can universally improve the recognition performance of the classifiers provided that the correct set of features are identified. The ability of a feature to contribute towards the identification of a pattern as belonging to a class and as not belonging to another class plays the most important aspect for performance boosting of classifiers. In this paper, we present a method to select the features from high dimensional training data by studying the individual ability of a feature to discriminate within and between classes by taking into account the area of overlaps resulting from the statistics of training data.

## 2 NEAREST ATTRIBUTES BASED DISCRIMINANT FEATURE SELECTION

The robustness and accuracy of pattern classification method is dependent on the availability of features in patterns that are most discriminating among the classes. Classical feature selection methods focuses on detecting the features based on strength of a feature within a database. The strong correlation that exists between the class discriminating ability of a classifier and the selected features is ignored in the conventional process of feature selection. The presented method recognizes the need for selecting features based on its discriminating nature among different classes by individually selecting features from nearest set of attributes that best represent the class samples among all the classes. Here, the nearest attributes are determined by assessing the area of overlap that results with inter-class and intra-class distance distributions for each attribute within a class. This section addresses how this area of overlap is calculated and how the nearest attributes are determined by an example application to a binary pattern classification problem.

### 2.1 Methodology and approach

To explain the proposed approach consider Wisconsin breast cancer database [32]–[34] having two classes Benign (total 458 samples) and Malignant (total 241 samples). Each sample in the database have 9 attributes that are: (1) clump thickness, (2) uniformity of cell size, (3) uniformity of cell shape, (4) marginal adhesion, (5) single epithelial cell size, (6) bare nuclei, (7) bland chromatin, (8) normal nucleoli and (9) mitoses. The feature values in these samples are between 1 and 10. This dataset is used for classification experiment by partitioning the data into two sets, training set and test set. These sets have equal number of samples selected randomly so that Benign class contains 229 train samples and Malignant class contain 120 train samples. Only training samples will be used in the process of detecting the most useful attributes in our approach to feature selection and classification.

In order to find the relative discriminating ability of the attributes in a class we look into overlap area of the distribution of differences from inter-class and intra-class feature comparisons. To illustrate this process, consider the attribute clump thickness that will have a total of 229 feature values from n = 229 training samples in Benign class and 120 feature values from 120 training samples in Malignant class. Let us denote feature values as $\mathbf{f_{i1}}(c_t)$ and $\mathbf{f_{i2}}(c_t)$, where i1 = {1...229} correspond to Benign class, i2 = {1...120} correspond to Malignant class and $c_t$ denotes clump thickness attribute. The intra-class ($d_{ia}$) and inter-class ($d_{ie}$) differences are those absolute valued differences between feature values within a given class and between classes respectively. In formal terms, for clump thickness attribute, $d_{ia}(c_t)$ is a set of differences $\{|\mathbf{f_i}(c_t) - \mathbf{f_j}(c_t)|\}, \forall i \in i1, j \in i1$ and $d_{ie}(c_t)$ is a set of differences $\{|\mathbf{f_i}(c_t) - \mathbf{f_j}(c_t)|\}, \forall i \in i1, j \in i2$. The intra-class differences for Benign and Malignant class would result in $\sum_{i=1}^{228}(229-i)$ values and $\sum_{i=1}^{119}(120-i)$ values respectively. The inter-class differences in Benign and Malignant class would result in $229 \times 120$ values and $120 \times 229$ values, respectively. Training and test sets are formed form this database by randomly selecting approximately equal numbers of instances. The training set would contain 114 instances of Benign class and 60 instances of Malignant class, while the test set would have 115 instances of Benign class and 60 instances of Malignant class.

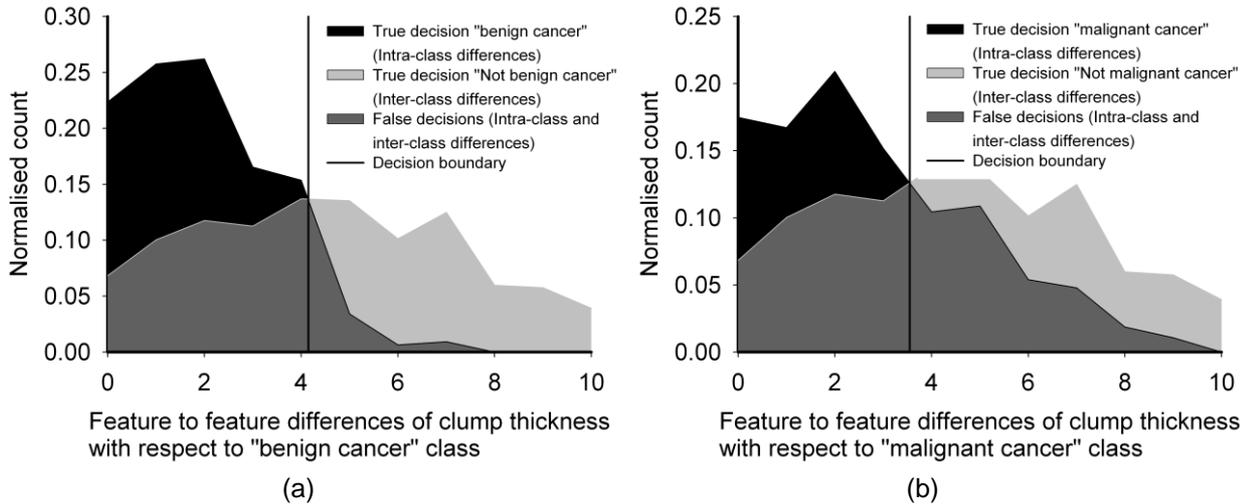

Fig. 1. Area normalized histograms of intra-class and inter-class differences in a two class classification problem generated from feature to feature comparison using the clump thickness attribute are shown. Using the intersection of intra-class and inter-class difference as the reference decision boundary, the plots (a) and (b) shows regions of correct detection representing true decisions and the region of confusion representing false decisions for Benign cancer class and Malignant cancer class respectively. Minimum number of false decisions occur when decision boundary is assigned at the intersection of inter-class and intra-class distance distributions. The area of this region of overlap between intra-class and inter-class differences gives the measure of minimum number of false decisions, and in turn shows the maximum discriminatory ability of an attribute for classifying a sample from a class.

The area normalized histograms of the inter-class and intra-class distance values for clump thickness attribute from the training set is shown in Fig. 1. Total area under each of the distance histograms is normalized to a value of 1 by dividing the histogram counts with total area under the histogram. Area of overlap between the inter-class and intra-class distances shows the natural discriminatory ability of an attribute belonging to a class. Figure 1(a) and 1(b) illustrate the overlap areas resulting from the inter-class and intra-class distance distribution with respect to the Benign class and Malignant class respectively. The decision boundary shown in Fig 1 partitions the histogram regions based on the intersection of intra-class and inter-class differences. The three distinct regions are shown, they are: (1) the region of true decision as belonging to a class, (2) the region of true decision as not belonging to a class, and (3) the combined region of overlap representing false decisions as belonging or not belonging to a class. The area of the overlap region represents a measure of false decisions (acceptance or rejection) of a sample belonging to a class when using clump thickness attribute alone for the task of classification. An area of overlap having a value equal to zero implies maximum feature discrimination, while a value of one would mean that the feature is not discriminable. From Fig 1, it can be seen that when using the clump thickness attribute, the area of overlap is more for Malignant class than Benign class, which means that when clump thickness alone is used for classfication the chance of accurately predicting the class of an unknown sample belonging to Benign class is more than a sample that belongs to a Malignant class.

To calculate the area of overlap, the region of overlap is divided into several trapezoids as shown in Fig 2. Referring to the X and Y coordinates depicted in Fig 2, the area of overlap ($A_o$) can be calculated as:

$$A_o(c_t) = \frac{1}{2} \sum_{i=1}^{\aleph} (X_{i+1} - X_i)(Y_{i+1} + Y_i) \qquad (1)$$

Applying Eq. (1) on Benign and Malignant classes for clump thickness ($c_t$) attribute shown in Fig 1 results in $A_o$ values of 0.42 for Benign and 0.76 for Malignant. Repeating this process of $A_o$ calculation for all the attributes we arrive at the values shown in Table 1. In Table 1, the relative area of overlap $A_r$ is calculated by subtracting the $A_o$ values with the minimum $A_o$ among all the attributes within a class. For the illustrative example both Benign and Malignant classes have the minimum value of $A_o$ for Uniformity of cell size attribute. The low values of $A_r$ shows the attributes that are most useful and large values of $A_r$ shows the attributes that are least useful in the process of classification. Obviously, the attributes represented by low values of $A_r$ need to be retained and selected for a classification task. The minimum values of $A_r$ among all the classes that is represented as $A_{min}$ in Table 1 are used as the measure to select the attributes.

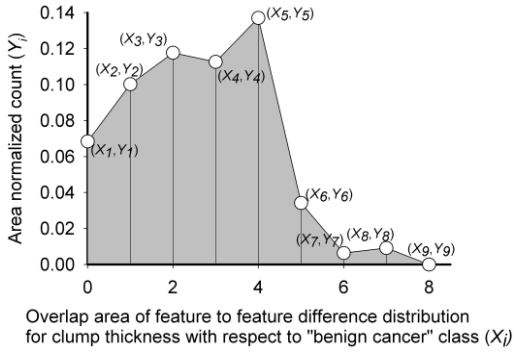

Fig. 2. Trapezoidal approximation to overlap area calculation for clump thinness attribute in Benign class.

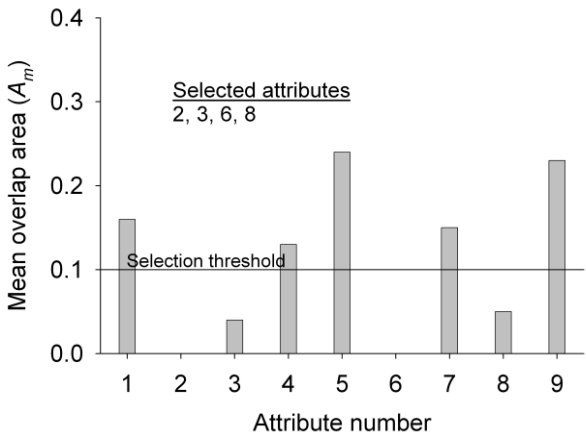

Fig. 3. The attributes having a relative overlap area less than a selection threshold is selected as most discriminating attributes. Four out of a possible nine attributes are selected by keeping the selection threshold at value of 0.1.

Figure 3 shows the nearest set of attributes that best represent the class that are within the tolerance of 10% of the minimum value (i.e. with $A_{min}$ value less than 0.1). For the example in Table 1, the minimum overlap area for the ith attribute $A_{min}(i)$:

$$A_{min}(i) = min(A_{r(i)}(Benign), A_{r(i)}(Malignant)) \quad (2)$$

The use of $A_{min}$ ensures that all the attributes that have its least value among different classes less than the selection threshold is selected. It can be seen from Fig 3 that the use of $A_{min}$ results in 4 attributes out of possible 9 getting selected. The effectiveness of the selection processes explained in Fig 3 can be verified using a classification experiment. The simplest approach to perform such classification is using a nearest neighbor classifier.

Table 2 show the selected attributes based on $A_{min}$ values from Table 1. Attributes that fall within a desired selection threshold is individually evaluated using leave one out cross-validation on gallery data with a nearest neighbor classifier. The attributes that have the highest recognition accuracy is ranked the highest. The combined accuracy shows the recognition accuracy on testing with the test set when using the combination of attributes. It can be seen that the use of top ranked attribute gives an accuracy of 94.00%, while the use of top 4 attributes give an accuracy of 95.43%. This is in contrast with an accuracy of 94.51% when all the attributes are used for classification without applying feature selection.

### 2.2 Heuristic determination of selection threshold

By default selection threshold of less than 0.2 would be sufficient when the variability between the areas of overlap between the attributes from each class will be less. On the other hand when the variability between the areas of overlap between the attributes from each class is large, it can be assumed that a higher value of threshold such as 0.3 would be required. The selection of the threshold can be automated by leave one out cross validation test on training set. The selection threshold that gives highest recognition accuracy is selected as the optimal value. Figure 4 shows the recognition accuracy results of cross validation for the optimal selection of threshold. Leave one out cross validation is performed on training set formed from 50% split of Wisconsin breast cancer database. It can be seen the best recognition performance is obtained around a threshold value of 0.2. Unless otherwise mentioned, we use this value of selection threshold to demonstrate the performance of the presented feature selection approach in the experimental verification sections of this paper.

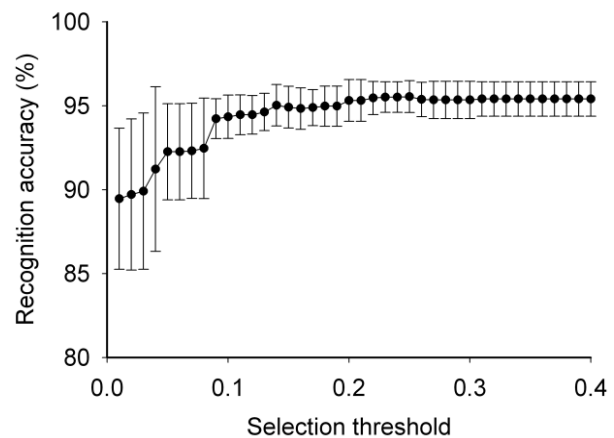

Fig. 4. Recognition performance of cross-validation test using the nearest neighbour classifier when using the presented approach of selecting features for Wisconsin breast cancer database [32]–[34].

5TABLE 1
Example of area of overlap and relative area of overlap in determining minimum overlap of attributes

| Attribute number ($A_i$) | Attribute | Benign | | Malignant | | |
|---|---|---|---|---|---|---|
| | | Overlap area $A_o$ | Relative Overlap area $A_r$ | Overlap area $A_o$ | Relative Overlap area $A_r$ | Minimum overlap $A_{min}$ |
| 1.00 | Clump thickness | 0.42 | 0.22 | 0.76 | 0.16 | 0.16 |
| 2.00 | Uniformity of cell size | 0.20 | 0.00 | 0.77 | 0.17 | 0.00 |
| 3.00 | Uniformity of cell shape | 0.25 | 0.04 | 0.72 | 0.12 | 0.04 |
| 4.00 | Marginal adhesion | 0.34 | 0.13 | 0.87 | 0.27 | 0.13 |
| 5.00 | Single epithelial cell size | 0.44 | 0.24 | 0.91 | 0.31 | 0.24 |
| 6.00 | Bare nuclei | 0.24 | 0.03 | 0.60 | 0.00 | 0.00 |
| 7.00 | Bland chromatin | 0.38 | 0.18 | 0.75 | 0.15 | 0.15 |
| 8.00 | Normal nucleoli | 0.25 | 0.05 | 0.81 | 0.21 | 0.05 |
| 9.00 | Mitoses | 0.44 | 0.23 | 0.85 | 0.25 | 0.23 |

TABLE 2
Performance demonstration of selected and ranked features

| Attribute number | Attribute name | Individual cross-validation accuracy (%) | $A_{min}$ | Combined accuracy (%) |
|---|---|---|---|---|
| 2.00 | Uniformity of Cell Size | 72.70 | 0.00 | 94.00 |
| 8.00 | Normal Nucleoli | 61.67 | 0.05 | 95.14 |
| 3.00 | Uniformity of Cell Shape | 60.62 | 0.04 | 95.14 |
| 6.00 | Bare Nuclei | 60.62 | 0.00 | 95.43 |

TABLE 3
Organisation of the databases used in the experiments

| Database | Number of instances | Number of features | Number of classes | Category |
|---|---|---|---|---|
| GLI-85 [35] | 85 | 22283 | 2 | Micro-array |
| GLA-BRA-180 [36] | 180 | 4915 | 4 | Micro-array |
| CLL-SUB-111 [37] | 111 | 11340 | 3 | Micro-array |
| TOX-171 [38] | 171 | 5748 | 4 | Micro-array |
| SMK-CAN-187 [39] | 187 | 19993 | 2 | Micro-array |
| AR100P [40] | 2600 | 2400 | 100 | Image |
| PIX10P [41] | 100 | 10000 | 10 | Image |
| PIE10P [42] | 210 | 2420 | 10 | Image |
| ORL10P [43] | 100 | 10304 | 10 | Image |

## 3 EXPERIMENTAL VERIFICATION

The classical use of methods for feature selection or dimensionality reduction of feature vectors have been to improve the recognition performance for classification problems involving high dimensional data. In this section we address this aspect through standard datasets used for benchmarking feature selection methods.

### 3.1 Micro-array data classification

Advancement in measurement techniques and computing methodologies has resulted in the use of micro-array data in various studies and application to genetics, medicine and diagnosis. The high dimensionality of the feature vectors in the mircoarray data often contains features that are not useful in the process of classification.

#### 3.1.1 Database organisation

3.1.1.1 GLI-85/GSE4412 [35]: Diffuse infiltrating gliomas are the most common primary brain malignancy is found in adults, and Glioblastoma multiforme, the highest grade glioma, is associated with a median survival of 7 months. The database is made from the transcriptional profiling of 85 gliomas from 74 patients to elucidate glioma biology, prognosticate survival, and define tumor sub-classes. The classification task of glioma tumor sub-classes is done through the database formed of large scale gene expressions. Two classes (grade III and IV gliomas) are used for the classification study as the distinction between these grades is most difficult. A total of 22283 gene expression attributes are used and in total there are 85 instances of gliomas.

3.1.1.2 GLA-BRA-180/GDS1962 [36]: GDS1962 dataset has its application in the analysis of analysis of gliomas of different grades. The database consists of the expression profile of Stem cell factor useful to determine tumor angiogenesis. The four classes in this database are (1) brainoligodendrogliomas, (2) glioblastomas, (3) astrocytomas and (4) non-tumor. There are 4915 attributes and 180 instances.

3.1.1.3 CLL-SUB-111/GSE2466 [37]: The database has gene expressions from high density oligonucleotide arrays containing genetically and clinically distinct subgroups of B-cell chronic lymphocytic leukemia (B-CLL). The dataset is formed of 11340 attributes and 111 instances.

3.1.1.4 TOX-171/GDS2261 [38]: This database is an example of the use of toxicology to integrate diverse biological data, such as clinical chemistry, expression, and other types of data. The database contains the profiles resulting from the three toxicants: alpha-naphthyl-isothiocyanate, dimethylnitrosamine, and N-methylformamide administered to rats. The classification task is to identify whether the samples are toxic, non toxic or control. Sample is toxic if alpha-naphthylisothiocyanate, or dimethyl-nitrosamine or n-methylformamide is administered, non-toxic if caerulein or dinitrophenol or rosiglitazone is administered and control if untreated.

3.1.1.5 SMK-CAN-187/GSE4115 [39]: The database consists of gene expression data from smokers with lung cancer and without lung cancer. This is diagnostic gene expression profile that could be used to distinguish between the two classes. The database consists of 19993 gene expression attributes and 187 instances.

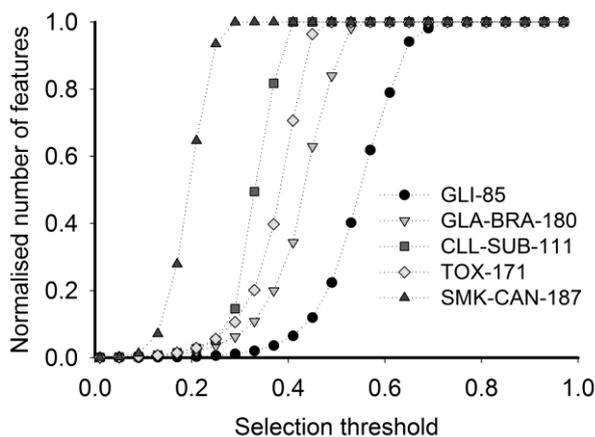

Fig. 5. The dependence of selection threshold on the number of selected features is shown for 5 gene expression databases.

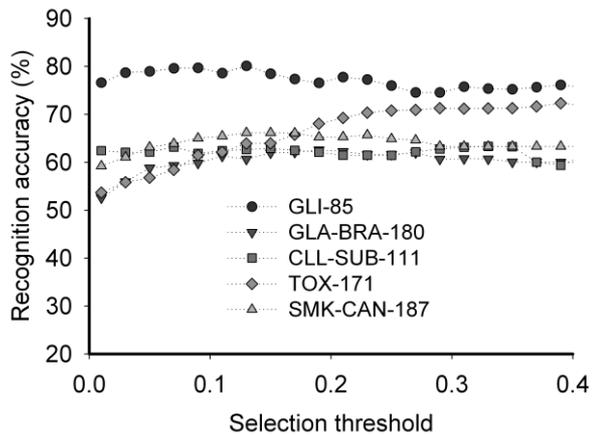

Fig. 6. Average recognition performance of the nearest neighbour classifier when using the presented approach of selecting features is shown for 5 gene expression databases.

### 3.1.2 Selection threshold and classification

In assessing the recognition performance of the presented feature selection method for micro-array databases listed in Table 3, we randomly select equal number of samples for forming the training and test set. It should be noted that for all the experiments and results presented in this section, a random split of 50% is done for individual classes in the databases to form the train and test sets. The average recognition accuracies are reported for 30 repeated random splits. The number of features that have an area of overlap within a specified selection threshold can vary from one database to another. This means that the quality of feature can vary from a database based on the the levels of natural variability within a database. Figure 5 shows this observation of quality of features as represented through the selection threshold and normalised number of features. It can be seen that for almost every database the quality of features varies when assessed purely based on the area of overlap. Interestingly, Fig. 5 shows that apart from SMK-CAN-187 database, the remaining databases contain only less than 3% of features relative to total number of features within a selection threshold of 0.2. This means that the intra-class variability in SMK-CAN-187 is lower than the other databases, and is possibility because lung cancer effects several gene expressions distinctively in comparisons with other cancer and toxicology databases.

Figure 6 shows the recognition performance of the presented feature selection method when used with nearest neighbor classifier. It can be seen that for all the databases, a selection threshold of 0.3 or lesser is sufficient to obtain high recognition accuracies. The maximum values of accuracies are possibly limited by the nature of the classifier and quality of the feature present.

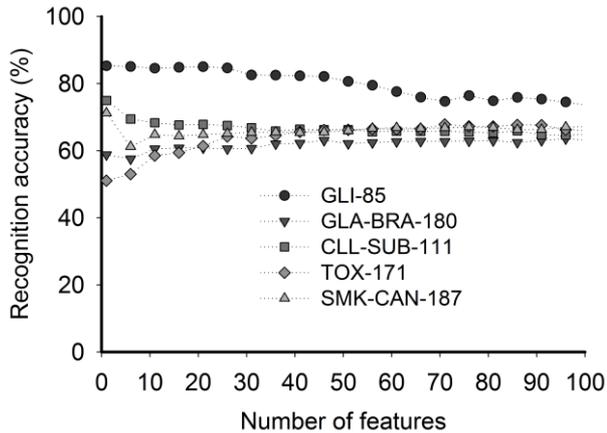

Fig. 7. Average recognition accuracies when using the top 100 selected attributes are shown. Nearest neighbor classifier is used to show the classification performance on 5 gene expression databases.

### 3.1.3 Feature ranking and classification

When the area of overlap between all the attributes is nearly same, applying threshold based selection results in the use of almost all available features for classification. The use of complete set of features in the process of automatic classification is often not a feasible option due to the issues of curse of dimensionality. In such situations, ranking the features and selecting any top ranked number of features can both serve the purpose of dimensionality reduction, and can also serve the purpose of selecting the best available features for classification. The simplest and common approaches to select the top ranks are by an individual search that evaluates each feature separately. Leave one out cross-validation is performed using the training set on individual features that are selected based on a specified value of selection threshold. The selected features are ranked based on least recognition error by evaluating it individually against a nearest neighbor classifier.

Figure 7 shows the average recognition accuracy of selecting top 100 features. These 100 features fall below the selection threshold of 0.2 and are ranked based on the least recognition error using the cross validation test. It can seen that by ranking the features and selecting the top features we group the features based on its discriminating ability. This result in increased recognition accuracies with very few numbers of features in all the databases.

### 3.1.4 Comparisons

Table 4 show the comparison of the best accuracies obtained with top ranked features using four conventional classifiers: nearest neighbor, linear SVM and naive Bayes. Overall, it can be seen that all the classifiers perform equally well. It should be noted here that in most cases, the highest recognition accuracies are obtained with very low number of features in comparison with the total number of available features. This means that for gene expression databases only very few gene expressions are useful for the process of classification irrespective of the type of classifier employed.

Table 5 shows the performance comparison of the presented feature selection method with that of conventional feature selection methods [44], [45]. It can be seen that the presented method uses fewer number of features to achieve higher recognition accuracy, which shows that the presented method results in more accurate selection of features useful for recognition than the conventional methods.

## 3.2 Image data classification

Face recognition is one of the challenging and most widely studied problems in recent years that contain large number of pixels as its attributes. In feature selection experiments, it is logical to consider each pixel as a feature and by applying the feature selection method the best set of pixels that is representative of the face can be used for classification. To benchmark the study on feature selection for face image recognition problem we select the subsets of AR [40], PIE [42], PIX [41] and ORL [43] posted on ASU feature selection summary website [41].

### 3.2.1 Database organisation

AR10 database is a subset of AR face database that is a benchmark database for feature selection studies. The main characteristic of this face database is that it contains images with occlusions, illumination and facial expression changes. The face images used in our experiments contain a total of 2400 pixels. The application of the mask to exclude background pixels results in 1778 in the face region as shown in Fig 8(a) and are used in our experiments for feature selection. PIX10 database is formed from the face images collected from the pilot European image processing archive. The PIX10 database contains face images with changing expressions and poses. The images in this contain a total of 10000 pixels. As shown Fig 8(b) the application mask to exclude the background results in 2939 pixels in the face region and is only used in the feature selection experiments. PIE10 database is a subset of PIE database that contain significant variation in illumination between the images and has a total of 2420 pixels. As can be seen in Fig 8(c) after the application of mask to remove background pixels a total of 1806 pixels remain in the face region. ORL10 is cropped face images from ORL database. This database contains large variation of poses between the images. After the mask is applied to remove the background pixels, only 6499 out of 10304 pixels are suitable to be used for feature selection task for face recognition as shown in Fig 8(d).

TABLE 4
The highest recognition accuracies of gene expression data when selecting features within the top 100 ranked features

| Database | Total number of features | Nearest neighbour | | SVM | | Naive bayes | |
|---|---|---|---|---|---|---|---|
| | | Accuracy (%) | Selected no. of features | Accuracy (%) | Selected no. of features | Accuracy (%) | Selected no. of features |
| GLI-85 | 22283 | 88.3 ± 2.9 | 3 | 86.5 ± 5.2 | 2 | 89.1 ± 2.9 | 3 |
| GLA-BRA-180 | 4915 | 65.3 ± 4.6 | 45 | 66.7 ± 4.8 | 6 | 68.4 ± 5.1 | 37 |
| CLL-SUB-111 | 11340 | 74.9 ± 2.6 | 1 | 65.6 ± 5.5 | 78 | 66.49 ± 8.37 | 50 |
| TOX-171 | 5748 | 69.6 ± 4.4 | 89 | 78.5 ± 5.5 | 71 | 61.5 ± 5.1 | 68 |
| SMK-CAN-187 | 19993 | 71.2 ± 1.7 | 1 | 73.2 ± 3.2 | 48 | 70.8 ± 4.0 | 52 |

TABLE 5
Comparison of top recognition accuracies using top 100 ranked features using a nearest neighbour classifier

| Database | Total number of features | MRMR [44] | | Information gain [45] | | Presented | |
|---|---|---|---|---|---|---|---|
| | | Accuracy (%) | Selected no. of features | Accuracy (%) | Selected no. of features | Accuracy (%) | Selected no. of features |
| CLL-SUB-111 | 11340 | 64.5 ± 6.7 | 32 | 64.2 ± 8.0 | 34 | **74.9±2.6** | 1 |
| SMK-CAN-187 | 19993 | 65.1 ± 4.3 | 41 | 65.1 ± 3.8 | 29 | **71.2±1.7** | 1 |
| GLI-85 | 22283 | 83.4 ± 4.5 | 67 | 84.2 ± 5.0 | 87 | **88.3±2.9** | 3 |
| GLA-BRA-180 | 4915 | 64.8 ± 3.4 | 45 | 65.6 ± 4.5 | 27 | **68.4±5.1** | 37 |
| TOX-171 | 5748 | 66.2 ± 5.1 | 100 | 65.5 ± 5.0 | 92 | **69.6±4.4** | 89 |

### 3.2.2 Selection threshold and classification

Figure 9 shows the dependence of numbers of features on the selection threshold for the four face image databases. Even with a low selection threshold of 0.2, it can be seen that in the face image databases about more than 20% of the features are selected. In comparison with micro-array datasets in Fig 5, it can be seen from Fig. 9 that image datasets have a larger percentage of good quality features that are useful for classification.

In conventional literature, the pixels in facial features such as eye, nose, and mouth are considered to have the most important information for recognition of a face and this idea has been used by researchers for face classification and identification problems. Since selected features (pixels) in images belong to facial features such as eye, nose etc, the location of these selected features in the facial region would indicate the importance of facial features in classification. The red color pixels in Figure 8 show the location of the features that are selected within a threshold of 0.2. From Fig. 8, a definitive inference on the importance of the location of a pixel being part of a face in a database to another is largely unclear. This would indicate that the discriminatory pixels in a face image can result from any part of the face image, and that it is not entirely depended on the face structure itself, but also on the type of natural variability that is inherent in a database.

Similar to the experiments using gene expression data, a random split of 50% is done for forming training and testing sets for all the databases, and the average recognition accuracies obtained by testing over 30 repeated random splits are presented. The recognition accuracy of the presented method when selecting different thresholds is shown in Figure 10. Like gene expression databases, in face image databases also a threshold of less than 0.3 is again sufficient to ensure high recognition accuracy.

### 3.2.3 Feature ranking and classification

The green color pixels shown in Figure 8 represents those features that are the top 100 pixels ranked according to its discriminating ability. These pixels are selected as the top rank based on leave one out cross validation on individual features that fall within the selection threshold of 0.2 in the training set using a nearest neighbor classifier. It can be seen from Fig. 8 that from a face database to the other, the importance of the pixels located in the face vary considerably. This is largely due to the different type of intra-class natural variability in each of these databases.

Figure 11 shows the average recognition accuracy of the top 100 features that falls within the selection threshold of 0.2. It can be seen from Fig. 11 that with as few as 100 features the recognition accuracies obtained are close to the highest accuracies observed in Fig. 10. This indicates the relative importance of the top ranked features shown in Fig. 8 (shown as green pixels) in the process of classfication.

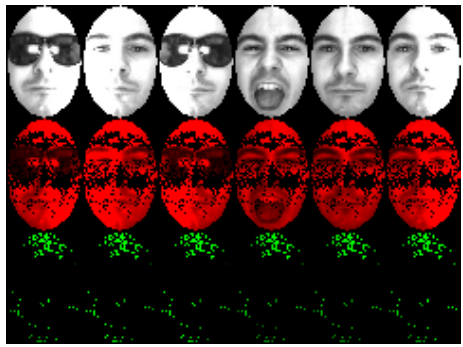

(a)

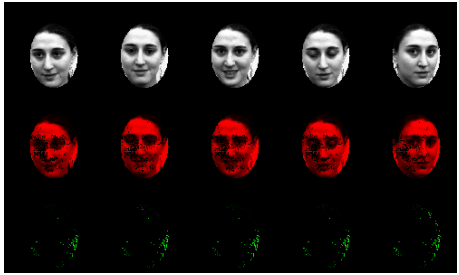

(b)

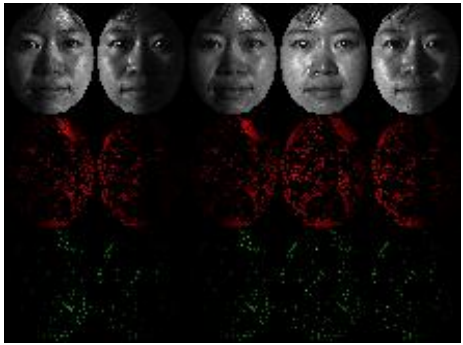

(c)

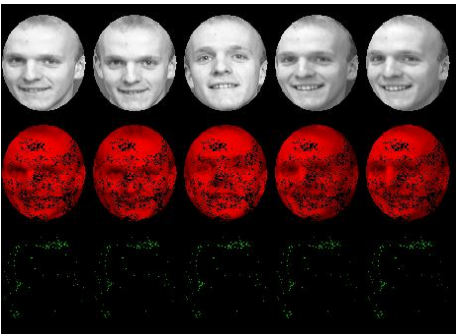

(d)

Fig. 8. An illustration of pixel in face images when a selection threshold of 0.2 is used for selecting features is shown using red color pixels. The 100 top ranked pixels among these selected pixels are shown in green color. The images (a)-(d) correspond to example training images from AR10, PIX10, PIE10 and ORL10 databases.

*3.2.4 Comparisons*

Table 6 shows the performance comparison of nearest neighbor, linear SVM and naive bays classifiers with respect to the highest recognition accuracies obtained by using the top ranked features. It can be seen that image data in comparison with gene expression data needs more number of features for achieving higher accuracies. This could be because most attributes in a face are important for its recognition, as opposed to a gene expression where the only few genes describe the occurrence of a cancer. Table 6 also show that with as few as 100 pixels that are selected based on its discriminatory ability good recognition accuracies can be achieved. Obviously, a higher recognition accuracy can be obtained with more number of pixels.

A comparison with other feature selection methods in Table 7 shows that the presented method performs better with respect to recognition accuracy although number of features that all the methods require is almost identical.

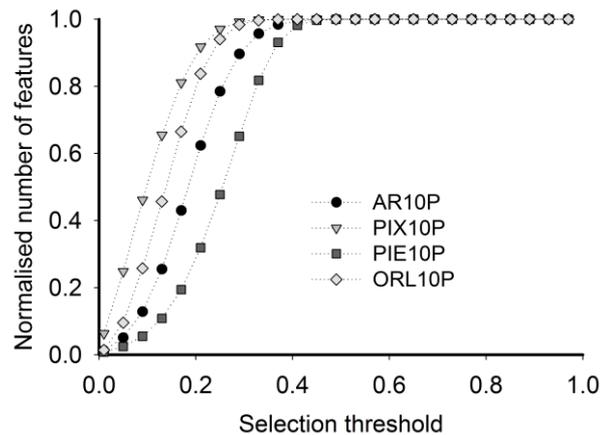

Fig. 9. The dependence of selection threshold on number of attributes for different face image databases is shown. The normalized number of attributes is the ratio between total number of selected attributes and that of total number of available attributes in a database.

## 4 CONCLUSION

The paper presented a feature selection method that is based on the assessment of discriminatory ability of individual features within a class. Area of overlap of inter-class and intra-class distance distribution of individual attributes is used as the main assessment criteria. The ability of the presented method to select most discriminatory features improves the performance recognition and it only requires fewer number of features in comparison with conventional methods. The number of features that are required for achieving high recognition accuracy varies from one database to another. A common framework to select the most important set of features is provided by applying selection threshold. The improved performance of using the presented method was demonstrated using gene expression data and face image data. The example applying the presented method on two very distinct high dimensional databases shows the general applicability of the method in other specific pattern recognition problems.

TABLE 6
The highest recognition accuracies of image data when selecting features within the top 100 ranked features

| Database | Total number of features considered | Nearest neighbour | | SVM | | Naive bayes | |
|---|---|---|---|---|---|---|---|
| | | Accuracy (%) | Selected no. of features | Accuracy (%) | Selected no. of features | Accuracy (%) | Selected no. of features |
| AR10P | 1778 | 79.2±5.4 | 32.0 | 83.3±5.6 | 91.0 | 65.6±3.8 | 70.0 |
| PIX10P | 2939 | 87.9±4.5 | 88.0 | 89.3±4.8 | 100.0 | 89.5±3.6 | 96.0 |
| PIE10P | 1806 | 89.2±4.0 | 85.0 | 96.8±2.6 | 71.0 | 73.2±7.9 | 95.0 |
| ORL10P | 6499 | 86.7±4.2 | 99.0 | 88.4±4.2 | 100.0 | 80.5±6.0 | 95.0 |

TABLE 7
Comparison of top recognition accuracies on image data using top 100 ranked features using a nearest neighbour classifier

| Database | Total number of features considered | MRMR [44] | | Information gain [45] | | Presented | |
|---|---|---|---|---|---|---|---|
| | | Accuracy (%) | Selected no. of features | Accuracy (%) | Selected no. of features | Accuracy (%) | Selected no. of features |
| AR10P | 1778 | 76.3±6.2 | 100.0 | 77.7±4.6 | 89.0 | **79.2±5.4** | 32.0 |
| PIX10P | 2939 | 81.3±5.9 | 100.0 | 85.4±6.0 | 84.0 | **87.9±4.5** | 88.0 |
| PIE10P | 1806 | 84.0±4.7 | 100.0 | 87.8±3.7 | 99.0 | **89.2±4.0** | 85.0 |
| ORL10P | 6499 | 71.3±4.5 | 100.0 | 78.3±5.1 | 97.0 | **86.7±4.2** | 99.0 |

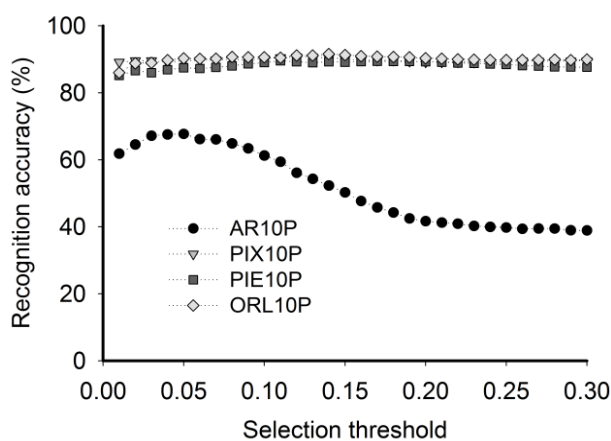

Fig. 10. Average recognition accuracies corresponding to various selection thresholds for face image datasets are shown.

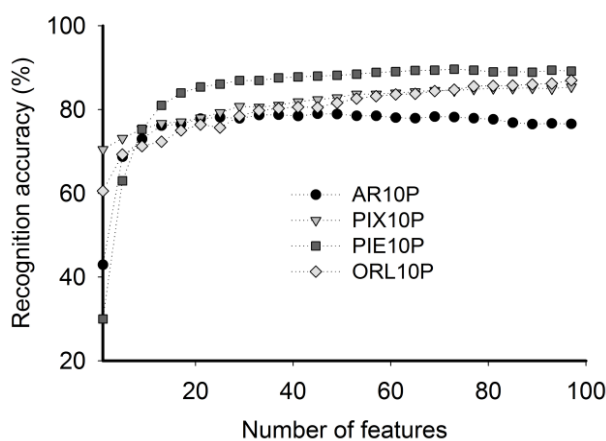

Fig. 11. Average recognition accuracies for face recognition problem when using the top 100 selected features are shown.